\documentclass[conference]{IEEEtran}
\IEEEoverridecommandlockouts
\usepackage{cite}
\usepackage{fancyhdr}
\usepackage{amsmath,amssymb,amsfonts}
\usepackage{algorithmic}
\usepackage{graphicx}
\usepackage{textcomp}
\usepackage{xcolor}
\usepackage{url}
\usepackage{balance}
\def\BibTeX{{\rm B\kern-.05em{\sc i\kern-.025em b}\kern-.08em
    T\kern-.1667em\lower.7ex\hbox{E}\kern-.125emX}}

\makeatletter
\newcommand{\linebreakand}{%
  \end{@IEEEauthorhalign}
  \hfill\mbox{}\par
  \mbox{}\hfill\begin{@IEEEauthorhalign}
}

\newcommand*\titleheader[1]{\gdef\@titleheader{#1}}
\AtBeginDocument{%
  \let\st@red@title\@title
  \def\@title{%
    \bgroup\normalfont\large\centering\@titleheader\par\egroup
    \vskip1.5em\st@red@title}
}
\makeatother

\title{Asch Meets HRI:\\Human Conformity to Robot Groups}
\titleheader{This paper has been accepted for presentation at the ``advancing GROup UNderstanding and robots' aDaptive behavior" (GROUND) workshop at the IEEE RO-MAN 2023 conference, Busan, South-Korea.}

\author{
\IEEEauthorblockN{Jasmin Bernotat}
\IEEEauthorblockA{CONTACT Unit \\
Italian Institute of Technology\\
Genoa, Italy \\
jasmin.bernotat@iit.it}
\and
\IEEEauthorblockN{Doreen Jirak}
\IEEEauthorblockA{CONTACT Unit \\
Italian Institute of Technology\\
Genoa, Italy \\
doreen.jirak@iit.it}
\and
\IEEEauthorblockN{Eduardo Benitez Sandoval}
\IEEEauthorblockA{Art and Design, Creative Robotics Lab (CRL) \\ 
UNSW Sydney \\
Sydney, Australia \\
e.sandoval@unsw.edu.au}
\linebreakand 
\IEEEauthorblockN{Francisco Cruz}
\IEEEauthorblockA{School of Computer Science and Engineering \\
UNSW Sydney \\
Sydney, Australia \\
f.cruz@unsw.edu.au}
\and
\IEEEauthorblockN{Alessandra Sciutti}
\IEEEauthorblockA{CONTACT Unit \\
Italian Institute of Technology\\
Genoa, Italy \\
alessandra.sciutti@iit.it}
}

\begin{document}
\maketitle

\begin{abstract}
We present a research outline that aims at investigating group dynamics and peer pressure in the context of industrial robots. Our research plan was motivated by the fact that industrial robots became already an integral part of human-robot co-working. However, industrial robots have been sparsely integrated into research on robot credibility, group dynamics, and potential users' tendency to follow a robot's indication. Therefore, we aim to transfer the classic Asch experiment (see \cite{Asch_51}) into HRI with industrial robots. More precisely, we will test to what extent participants follow a robot's response when confronted with a group (vs. individual) industrial robot arms (vs. human) peers who give a false response. We are interested in highlighting the effects of group size, perceived robot credibility, psychological stress, and peer pressure in the context of industrial robots. With the results of this research, we hope to highlight group dynamics that might underlie HRI in industrial settings in which numerous robots already work closely together with humans in shared environments.
\end{abstract}

\begin{IEEEkeywords}
social conformity, peer pressure, trust in HRI, robot credibility, and interaction design 
\end{IEEEkeywords}

\section{Introduction}
Humans' behavior and beliefs can change and even steered in opposite directions under the influence of their peer groups that may define their own social norms and expected behaviors. Such an alignment is usually termed \textit{conformity} which can be defined as \textit{``the process whereby people change their beliefs, attitudes, actions, or perceptions to more closely match those held by groups to which they belong or want to belong or by groups whose approval they desire"} \cite{Brit}.
In 1951, Salomon Asch \cite{Asch_51} introduced his seminal experiment studying the level of conformity disguised as a visual perception task. Given a certain line length, a participant among a group of others was asked to select the correct answer from three given lines. The essence of the experiment was that except the current participant, all others were confederates 
who indicated the wrong line in most of the trials. Although participants initially provided correct answers, they confirmed to the obviously incorrect responses of the confederates over time. Moreover, participants rather explained their ``failure" by doubting their own visual perception or cognitive abilities. Within the last decades, several replications of Asch's experiment were conducted \cite{Crutc55,Nicho85,Mori10} including those that failed to show evidence for Asch's findings \cite{Adams84,Lalan90}. However, it needs to be considered that the Asch experiment took place in the early 1950ties in conservative America. The selection bias for participants (only white young male students) and the low sample size might have affected Asch's findings. However, until today, Asch's experiment is powerful as it well demonstrates the impact of a group on individuals.

\section{Related Work}
Aspects of social dynamics and peer pressure have expanded the research area of social psychology and robotics in order to include humanoid-like and socially-behaving robots in human environments. 

Brandstetter et al. \cite{Brand_14} conducted a study with a group of humanoid NAO robots in two tasks, one representing the Asch line judgment experiment and one a verbal task (adding the past tense of a given verb list). Additionally, the authors separated their stimuli into ambiguous and non-ambiguous ones and provided a baseline experiment that did not include peers. The authors found conformity effects for the human peer group but not for the robot group. Furthermore, the level of conformity was higher in the ambiguous condition than in the non-ambiguous condition.

Salomon et al. \cite{Salom_18} set up a group of three MyKeepon robots in two experimental conditions: Either the robots answered before (test condition) or after (control) participants' response. The authors altered the Asch line task by choosing a digital version of the ``Dixit" game with no objective true or false answer. Participants' tendency to conform with the robot group was way higher in the experimental condition than in the control condition. Moreover, the results underpinned the role of trust by demonstrating that once the robots lost their credibility, participants no longer conformed to the robots' responses. 

Elson et al. \cite{Elson_22} reproduced Asch's 
experiment within the framework of trait-activation theory, i.e. how a specific situation influences a human's behavior. In an online study, the authors provided a user interface that simulated a video call. The originally proposed lines in three different lengths served as the visual stimuli. A group of seven NAO robots represented the confederates' group which verbalized the decision on the line length (A-C) prior to the human response. Additionally, the answer was displayed in an extra video panel. One session consisted of 18 trials and in total \textit{N} = 119 volunteers within an age range of 19-24 years participated in this study. The results revealed that 52.1\% of the participants did not conform to the confederates' responses compared to 26\% in the Asch experiment. Moreover, highest conformity rates appeared in the first half of the trials but dropped significantly in the other half of the trials. Finally, a correlation analysis of the Big Five personality traits revealed that 'openness' was the most critical factor leading to conformity behavior.

A study investigating the tendency for conformity in adults vs. children has been presented by Vollmer et al. \cite{Vollm18}. The experimental setup consisted of two conditions with \textit{N} = 20 participants each (control group vs. NAO robot peer group) for the children (age 7-9) and an additional condition (human peer group) for the adults. After presenting the visual stimuli on a TV screen, every participant (robot) gave a verbal answer. The evaluation of the two experiments revealed high normative behavior for adult participants in the human peer group condition, reproducing results from the original Asch experiment. The same effects could not be shown for the robotic peer group. However, the evaluation of the second experiment revealed normative behavior for the children supported by a significant performance drop in the line judgment when exposed to the robotic group (critical trial). As a follow-up on this study, Qin et al. \cite{Qin22} demonstrated that when only one (NAO) robot was present, i.e. as a minority, also adults showed a tendency to conform with both the human-peer and robot-peer condition. 

Other studies investigating conformity deviated from the setup of the original Asch experiment to include effects on risk-taking behaviors \cite{Hanoc_21} (BART task), group norms \cite{Fuse_20} (quiz), and communication \cite{Volan_19} (robot search task). However, in the present study, we preserved the original Asch experiment \cite{Asch_51} but extend the conformity research to human-robot interaction with industrial robots. The study aims to show the effects of social attribution and thus influence on trust and, consequently, conformity in the absence of human physical appearance in contrast to the studies previously outlined using well-known humanoid robotic platforms. The results have practical relevance in industrial settings with critical safety and security measures for human-robot collaborations. Furthermore, non-humanoid robots are expected to take a relevant role in everyday future human-robot interactions rather than humanoid robots. Hence, exploring embodiments different from humanoid robots fills a relevant research gap.

For our experiment, we set up the following research questions and hypotheses:
\paragraph{Research Questions}
\begin{enumerate}
    \item Willingness to follow a robot's response: Do participants follow industrial robot vs. human confederates' responses?
    \item Number of robots: To what extent does the number of confederates (group of three vs. single) cause participants to follow others' responses?
    \item Psychological factors: What psychological processes might underlie participants' response behavior during the experiment?
\end{enumerate}

\paragraph{Hypotheses}
\begin{enumerate}
\item Participants conform to the confederate(s)' responses when confronted with a
    \begin{itemize}
        \parindent40mm \item [a)] robot
        \parindent40mm \item [b)] human. 
    \end{itemize}
\item The effects hypothesized in Hypotheses 1a and 1b are expected to be driven by the group size of three (vs. one) confederates which is expected to increase perceived
\begin{itemize}
        \item [a)] social pressure
        \item [b)] psychological stress.
\end{itemize}
\end{enumerate}
as suggested by Asch's classic experiments (see \cite{Asch_51}).

At the same time, we will test whether participants'
\begin{enumerate}
        \item [a)] Acceptance of the confederates' responses
        \item [b)] Cognitive vs. affective trust in the confederates
        \item [c)] Tendency to follow a group
        \item [d)] Social desirability
    \end{enumerate}

In robot conditions also:
\begin{enumerate}
        \item [e)] Positive vs. negative attitudes toward robots
        \item [f)] Experience with technology in general
        \item [g)] Experience with robots in particular
    \end{enumerate}

affect their conformity with the confederates' responses. The effects of these covariates will be controlled because participants' social desirability, attitudes toward robots, and experience with technology and robots have been found to affect participants' responses in previous HRI experiments (see e.g. \cite{Berno17a, Berno17b, Berno17c, Schif16}). 

 \begin{figure}[h]
 \centering
 \begin{minipage}{.5\textwidth}
 \centering
 \includegraphics[width=0.6\textwidth]{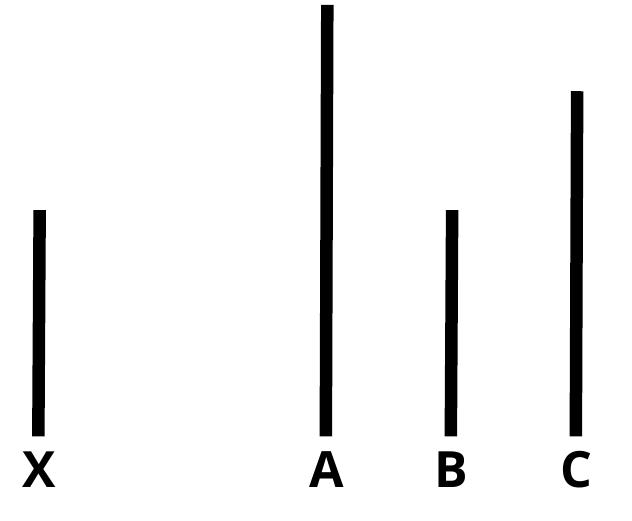}
 \end{minipage}%
 \hfill
 \begin{minipage}{.5\textwidth}
 \centering
 \includegraphics[width=0.6\textwidth]{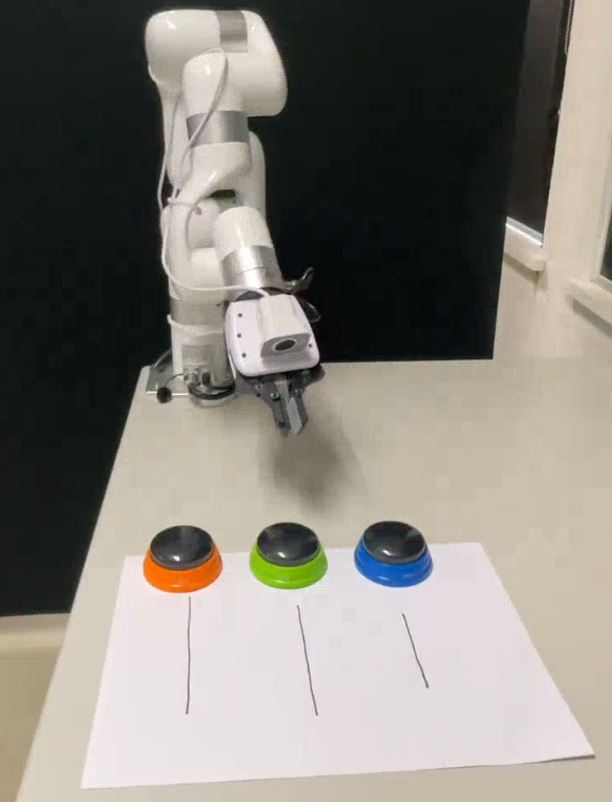} 
 \end{minipage}
  \hfill
 \begin{minipage}{.5\textwidth}
 \centering
 \includegraphics[width=0.6\textwidth]{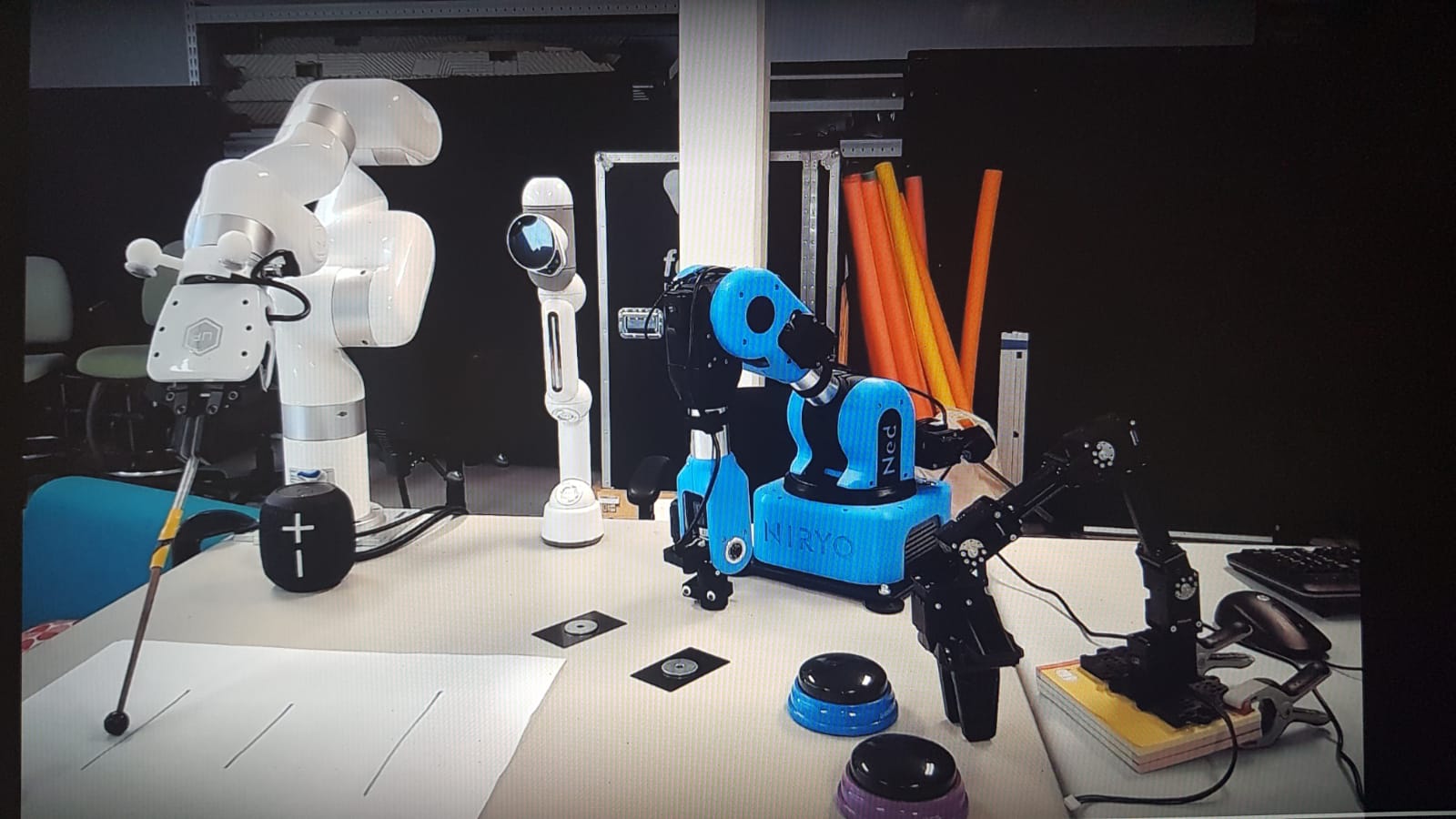}
 \end{minipage}
 \caption{Top: Example stimuli for the Asch experiment where 'X' is the reference line and lines 'A-C' the possible answers. Middle: A single robot showing the answer in a preliminary implementation. Bottom: Our setup of the Asch experiment displaying the robot group condition in a first prototype for the experiment (see section \ref{experimental_design})}
 \label{fig:Classical_Asch_setup}
 \end{figure}
 
\section{Experimental Design}
\label{experimental_design}
For the study, we propose to use a 2$\times$confederate type (robot vs. human)$\times$2 group size (1 vs. 3) between-subjects design.
The study will be conducted online via SoSciSurvey\footnote{\url{https://www.soscisurvey.de/}}. The link to participate will be spread on social media platforms, email lists, and universities in Italian and English-speaking countries. Following classic versions of the Asch experiment \cite{Asch_51}, the participants' task will be to judge the length of a line according to a reference line in several trials and to complete a short questionnaire. 

After having agreed to participate in the study with their data being recorded and used for scientific purposes (the study will be approved by the ethics committee of the Italian Institute of Technology, IIT), participants will first do some practice trials. These serve to make sure that participants are capable of judging the lengths of the lines correctly when being alone. Afterwards, participants will be introduced to the confederate(s) who will ostensibly already be waiting online for the participants to join.

Having initiated the study, a trial proceeds as follows: Four lines will be presented on the screen; one line, the reference line, will be left to three lines of different lengths (see Figure \ref{fig:Classical_Asch_setup}). The differences in lengths between the remaining two lines and the reference line will be no larger than 30\% \cite{Asch_51, Brand_14}. This is supposed to give room for reasonable doubt while the correct answer will be still obvious. Then, confederates and participants will be told in written form to indicate their responses one after another from left to right in the order in which their videos will br shown on the screen. As the confederates will be presented left to the participants, participants will respond last. Responses will be given by pressing the letter on the keyboard that refers to the respective line. Each given response will be displayed on the screen so that it will be visible to all attendees until the next attendee responds. In total, there will be 18 trials. In twelve trials the confederates (critical trials) will give the wrong response, while in six trials, the correct line will be indicated. Finally, participants will complete a questionnaire whose measures will be assessed in the same order as listed in section \ref{questionnaire_measures}. 
More specifically, we will use the following measures:

\subsubsection{Dependent Measures}
\begin{itemize} 
    \item Behavioral measure: Identification of the correct line
    \end{itemize}

\subsubsection{Questionnaire measures}
\label{questionnaire_measures}
    \begin{itemize}
        \item Perceived social pressure
        \item Perceived social stress
    \end{itemize}

\subsubsection{Covariates}
\begin{itemize}
    \item Acceptance of the confederate('s/s') responses
    \item Cognitive vs. affective trust in the confederate(s)
    \item Tendency to follow a group/individuals
    \item Social desirability \cite{Stoeb01}
    \end{itemize}

Additional covariates in human-robot interaction:
\begin{itemize}
    \item Positive vs. negative attitudes toward robots
    \item Experience with technology in general
    \item Experience with robots in particular
\end{itemize}

\subsubsection{Experience during the experiment}
\begin{itemize}
    \item Indication of how difficult it was for participants to identify the correct line
    \item Indication to what extent participants perceived the other confederates as a group
    \item Assessment of whether participants felt belonging to the group
    \item To what extent participants had adapted their responses to those of the confederate(s)    
\end{itemize}

\subsubsection{Manipulation Check} 
\begin{itemize}
    \item Test run without an interaction partner to identify whether they can differentiate between lengths
    \item Specification of the confederates as a) humans or b) robots. In case of robot confederates, also the specification of the robotic appearance as a) industrial, b) human-like, or c) zoo-morph
    \item Specification of the number of other participants involved
    \item Indication to what extent participants believed that they were really interacting with the confederates online
    \item \textbf{Only in robot conditions:} Indication whether participants had doubts the robot responded autonomously based on its technical abilities to perceive objects
    \item Enquiry of participants guesses about the purpose of the experiment
    \item Familiarity with the Asch experiment
\end{itemize}

\section{Contribution}
The human tendency to show conformity in the presence of a majority that might even go against one's own better knowledge or even beliefs and morals is still an important topic in social psychology as people are prone to manipulation which can, in the worst case, harm themselves and others. Especially in these days where robots immerse in the human environment, the understanding of normative conformity towards them either as a single entity \cite{Qin22} or in robot groups \cite{Vollm18, Wulle20} is of crucial significance. This is particularly true for the industrial context in which numerous robots work collaboratively with humans who have to trust the robot's technical abilities (e.g. controlling its force for collision avoidance) but also cognitive skills as in decision-making. Therefore, potential users' willingness to follow robots' advice might be decisive for a successful outcome of HRI. The overestimation of robots' functions, as well as a lack of robots' credibility and trust, might both lead to severe consequences \cite{Alhaj21, Choi21}. At the same time, HRI has increasingly been discovered as having huge similarities to interpersonal social processes amongst humans, such as group dynamics \cite{Eysse12, Eysse13}. In most studies, however, social robots were used to investigate group dynamics in HRI. Given that industrial robots already became an integral part of factory workers' everyday lives \cite{IFR23}, investigating the effects of group size, perceived robot credibility, psychological stress, and peer pressure in the context of industrial robots can guide directives to implement cognitive safety measures for human-robot collaborations.
With the reproduction of the classical Asch experiment on human conformity with industrial-like robot arms, we aim at studying the potential trust but also dangers in human-robot collaborations. We hope that our study hypotheses and experimental design stimulates other researchers in the field of social robotics, social psychology, and human-robot interaction for further analysis and discussion to foster evidence of trust between humans and robots.

\section*{Acknowledgment}
This work has been supported by a Starting Grant from the European Research Council (ERC) under the European Union’s Horizon 2020 research and innovation programme. G.A. No 804388, wHiSPER

\bibliographystyle{IEEEtran}
\balance
\bibliography{Asch_WS_GROUND2023_WithHeader.bib}

\end{document}